\documentclass[a4paper]{article}

\usepackage{authblk}
\usepackage{algorithm}
\usepackage{algorithmic}
\usepackage{graphicx}
\usepackage{amsmath}
\usepackage{amssymb}
\usepackage{amsfonts}
\usepackage{booktabs}
\usepackage{multirow}
\usepackage{amsthm}
\usepackage{microtype}
\usepackage{url}

\newcommand{\F}{\mathbb{F}}
\newcommand{\N}{\mathbb{N}}
\newcommand{\Z}{\mathbb{Z}}

\newtheorem{problem}{Problem}
\newtheorem{definition}{Definition}

\providecommand{\keywords}[1]{\textbf{\textit{Keywords }} #1}

\DeclareMathOperator{\fit}{fit}

\begin{document}

\title{Balanced Crossover Operators in Genetic Algorithms}

\author[1]{Luca Manzoni}
\author[2]{Luca Mariot}
\author[3]{Eva Tuba}

\affil[1]{{\normalsize Dipartimento di Matematica e Geoscienze, Università degli Studi di Trieste, Via Valerio 12/1, 34127 Trieste, Italy} \\

{\small \texttt{lmanzoni@units.it}}}

\affil[2]{{\normalsize DISCo, Universit\`{a} degli Studi di Milano-Bicocca, Viale Sarca 336/14, 20126 Milano, Italy} \\

  {\small \texttt{luca.mariot@unimib.it}}}

\affil[3]{{\normalsize Faculty of Informatics and Computing, Singidunum University Danijelova 32, 11000 Belgrade, Serbia} \\
  
  {\small \texttt{etuba@ieee.org}}}

\maketitle

\begin{abstract}
  In several combinatorial optimization problems arising in cryptography and
  design theory, the admissible solutions must often satisfy a balancedness
  constraint, such as being represented by bitstrings with a fixed number of
  ones. For this reason, several works in the literature tackling these
  optimization problems with Genetic Algorithms (GA) introduced new balanced
  crossover operators which ensure that the offspring has the same balancedness
  characteristics of the parents. However, the use of such operators has never
  been thoroughly motivated, except for some generic considerations about search
  space reduction.

  In this paper, we undertake a rigorous statistical investigation on the effect
  of balanced and unbalanced crossover operators against three optimization
  problems from the area of cryptography and coding theory: nonlinear balanced
  Boolean functions, binary Orthogonal Arrays (OA) and bent functions. In
  particular, we consider three different balanced crossover operators (each
  with two variants: ``left-to-right'' and ``shuffled''), two of which have
  never been published before, and compare their performances with classic
  one-point crossover. We are able to confirm that the balanced crossover
  operators performs better than all three balanced crossover operators.
  Furthermore, in two out of three crossovers, the ``left-to-right'' version
  performs better than the ``shuffled'' version.
\end{abstract}

\keywords{genetic algorithms, crossover operators, balanced bitstrings, Boolean functions, orthogonal arrays, bent functions}

\section{Introduction}
\label{sec:intro}
\emph{Crossover} (or \emph{recombination}) operators play a crucial role in
Genetic Algorithms (GA). The idea underlying crossover, borrowed from biological
evolution, is quite simple: given two candidate solutions, combining parts of
their chromosomes will yield an offspring potentially having better fitness than
the parents. This strategy stands on the observation that fit individuals share
some traits encoded at the chromosome level, which can be inherited by their
children via crossover. Indeed, this intuition has been formalized by
Holland~\cite{holland75} with the concept of \emph{building blocks} used in
\emph{schema theory}.

There exist several classes of combinatorial optimization problems whose
feasible solutions must contain a specified number of ones, i.e. they must have
a \emph{fixed Hamming weight}. Examples of such problems come, for instance,
from the domain of \emph{cryptography}, where \emph{balanced Boolean functions}
are used to design symmetric key cryptosystems~\cite{carlet-bool}. Another
research area where balanced binary strings are sought is that of
\emph{combinatorial designs}: there, one is interested in constructing subsets
of a certain support space (called \emph{blocks}) which satisfy specific
balancedness constraints~\cite{stinson}. A third research field where
fixed-weight bitstrings are used is that of \emph{portfolio optimization};
indeed, a portfolio can be represented by a binary vector where the positions
set to $1$ indicate that the corresponding assets have been
selected~\cite{kellerer}.

Genetic algorithms seem like a sensible choice for solving the optimization
problems mentioned above. However, breeding feasible solutions which have a
fixed Hamming weight is something that classic GA recombination operators such
as \emph{one-point} crossover cannot handle. As a matter of fact, starting from
two individuals with the same number of ones and applying one-point crossover
will likely produce an offspring having a different Hamming weight. This is due
to the fact that one-point crossover (as well as most other recombination
operators in the literature) does not enforce any control over the
multiplicities of the alleles copied in the offspring. Of course, this drawback
in dealing with fixed-weight bitstrings can be addressed at the fitness function
level. Since we do not have any guarantee that the offspring has the desired
number of ones, the idea is to add a \emph{penalty factor} to the fitness
function which punishes deviations from the expected Hamming weight. Although
being the simplest solution to cope with this constraint, one might argue that
it wastes a lot of fitness evaluations, because most of the solutions generated
by classic crossover operators will violate the fixed-weight property.

An alternative way to address this problem is to design new recombination
operators that \emph{preserve} the Hamming weight of the bitstrings, which we
term \emph{balanced crossover operators} in what follows. The first researchers
who pioneered this approach in the area of cryptography were Millan et
al.~\cite{millan98}, who proposed a counter-based crossover operator to evolve
balanced Boolean functions, which was later adapted to evolve \emph{plateaued
  functions} in~\cite{mariot15}. Similar operators have been later proposed for
GA applied to combinatorial designs problems~\cite{mariot17,mariot18}, portfolio
optimization~\cite{chen06, chen09}, multiobjective $k$-subset
selection~\cite{meinl09} and disease classification~\cite{sachnev16,sachnev18}.

Looking at the existing literature, one can remark that the introduction of
balanced crossover operators has never been thoroughly motivated. In fact, the
only recurring motivation supporting the use of such operators is the reduction
of the search space (see e.g.~\cite{mariot18}). To be sure, restraining the
crossover operator to produce only fixed Hamming weight bitstrings greatly
shrinks the space of candidate solutions searched by GA. Nonetheless, although
the reduction is quite evident for short strings, this advantage becomes less
clear as the Hamming weight $k$ approaches $n/2$, since
$\binom{n}{n/2} = \Theta(\frac{2^n}{\sqrt{n}})$. Moreover, most of the works in
the literature employing balanced crossover operators do not perform a sound
comparison of their results with those that can be obtained with classic
operators. Hence, it is not even clear on a statistical basis whether balanced
crossover operators actually bring any advantage to GA working with fixed
Hamming weight bitstrings.

The aim of this paper is to begin closing this gap by performing a thorough
statistical comparison of balanced and classic crossover operators over a set of
problems from the area of cryptography and design theory. In particular, we
consider three balanced crossover operators in our investigation: the first is a
modification of the counter-based operator proposed by Millan et
al.~\cite{millan98}. The other two, as far as our knowledge goes, have never
been published before, and they are based respectively on the \emph{map of ones}
and \emph{zero lengths} chromosome encodings. For all three crossover operators,
we have defined two variants: a ``left-to-right'' one, where the crossover is
applied as usual, and a ``shuffled'' version, where the positions of an
individual are randomly shuffled, the crossover is performed, and the positions
are shuffled back in order. This operations should, in principle, counterbalance
any positional bias in the crossover operator. As a term for comparison we
considered one-point crossover, optimizing the Hamming weight as a penalty
factor in the fitness function.

We considered three combinatorial optimization problems in our statistical
investigation. The first one regards \emph{nonlinear balanced Boolean
  functions}, where the goal is to maximize the nonlinearity of the functions
while retaining their balancedness. The second problem, always concerning
Boolean functions, is the evolution of \emph{bent functions}, which reach the
highest possible nonlinearity and, although unbalanced, have a specified Hamming
weight. Finally, the third problem pertains binary \emph{Orthogonal Arrays}
(OA), which are Boolean matrices having balanced subsets of columns.

We carried out our experiments over three different instances for each
problem. In order to compare the performances of the four crossover operators,
we employed a \emph{non-parametric} statistical test over the best individual
produced by each experimental run, namely the \emph{Mann-Whitney-Wilcoxon
  test}~\cite{garcia09}. Some works in the literature~\cite{picek11,picek13}
performed a comparison with non-parametric tests on classic crossover operators,
but did not consider balanced operators.

In our experiments we considered two main types of research questions: whether
the balanced operators perform better than the one-point crossover, and whether
there is any difference between the ``left-to-right'' and ``shuffled'' versions
of the operators. For the first question, we are able to answer affirmatively:
balanced crossover operators performs better, in general, than one point
crossover. In particular, using one point crossover seems to be a bad choice for
those problems where the balancedness constraint plays an important role in the
fitness function, i.e. in the OA problem. As for the second question, the answer
is more nuanced: for two out of three crossovers, removing any positional bias
is actually detrimental to the performances in some cases. Similarly to the
first question, this fact is more evident in the OA problem where balancedness
is a key component of the fitness function. Concerning the \emph{map of ones}
crossover, however, both the ``left-to-right'' and ``shuffled'' versions have
similar performances.

Although this work focuses on investigating the properties of balanced crossover
operators, rather than proposing them to outperform other state-of-the-art
evolutionary algorithms for problems in cryptography and combinatorial design
theory, we compare our results with some recent work on the subject,
namely~\cite{picek17} for balanced nonlinear and bent Boolean functions
and~\cite{mariot18} for orthogonal arrays. While in the first case our map of
ones crossover has a performance similar to those of the algorithms studied
in~\cite{picek17}, it still falls short of achieving the success rate of Genetic
Programming (GP) over the OA problem in~\cite{mariot18}, altough it improves on
the GA used in the same paper.

This work is an extended version of the short paper~\cite{manzoni19} presented
at GECCO 2019. In particular, the new contributions concern the experiments over
the bent functions and orthogonal arrays problems, and the comparison between
the ``left-to-right'' and ``shuffled'' versions of the balanced crossover
operators.

The remainder of this paper is structured as
follows. Section~\ref{sec:crossover} covers some basic background definitions
about fixed Hamming weight bitstrings, and describes in detail the three
balanced crossover operators investigated in our
study. Section~\ref{sec:problems} formally states the three optimization
problems considered in our investigation. Section~\ref{sec:exp-set} describes
the experimental design of our study, discussing in particular the structure of
the steady state GA employed in our experiments, and stating three research
hypotheses about the performances of balanced crossover
operators. Section~\ref{sec:res} presents the results of our experiments,
analyzing the performances of the considered crossover operators through
non-parametric tests. Section~\ref{sec:disc} engages in a discussion of the
three hypotheses in light of our experimental findings, and compares the
obtained results with some recent work in the literature. Finally,
Section~\ref{sec:conc} concludes the paper, and sketches some possible future
directions of research on the subject.

\section{Balanced Crossover Operators}
\label{sec:crossover}
In this section, we describe the three balanced crossover operators analyzed in
our experiments. Before delving into the details of each operator, we recall
some basic definitions and results about bitstrings and their Hamming weights.

Let $\F_2=\{0,1\}$ be the finite field with two elements. A \emph{bitstring} of
length $n \in \N$ is a binary vector $x$ of $n$ components, each of them
belonging to $\F_2$. We denote by $\F_2^n$ the set of all bitstrings of length
$n$. In what follows, we will often endow $\F_2^n$ with a vector space
structure, with bitwise XOR (denoted as $\oplus$) as vector sum and logical AND
as multiplication by a scalar from $\F_2$. Given a bitstring $x \in \F_2^n$, let
$supp(x) = \{i: x_i \neq 0\}$ be the \emph{support} of $x$, that is, the set of
coordinates equal to $1$ of the bitstring. The \emph{Hamming weight} $w_H(x)$ of
$x$ is then defined as the cardinality of its support, i.e.
$w_H(x) = |supp(x)|$. If $n$ is even and $w_H(x) = n/2$, we say that the
bitstring $x$ is \emph{balanced}. In other words, $x$ is balanced when it is
composed of an equal number of zeros and ones.

A binary string $x \in \F_2^n$ can be interpreted as the \emph{characteristic
  function} of a set $S \subseteq [n]=\{1,\cdots,n\}$. In particular, the
support of $x$ corresponds exactly to $S$, i.e. to the image of its
characteristic function. Basic combinatorial arguments show that the number of
all bitstrings of length $n$ is $|\F_2^n| = 2^n$, that is, the cardinality of
the power set $\mathcal{P}([n])$ of $[n]$. Likewise, for $k \in [n]$ the size of
the set $\mathcal{B}_{n,k}$ of bitstrings having Hamming weight $k$, or
equivalently the number of $k$-subsets of $[n]$, is $\binom{n}{k}$, since it
corresponds to the number of ways one can choose $k$ objects out of $n$.

The search space of interest for our investigation is precisely
$\mathcal{B}_{n,k}$, which we will also call the set of $(n,k)$-combinations in
what follows. In particular, $\mathcal{B}_{n,k}$ will represent the set of
feasible solutions to a particular optimization problem explored by GA. Although
several types of crossover operators have been proposed in the literature of GA,
very few of them consider restrictions on the Hamming weights of the
chromosomes, i.e. which actually restrict the search of a GA to
$\mathcal{B}_{n,k}$. More generally, such crossover operators are not
specifically designed to evolve $(n,k)$-combinations.

We now describe the crossover operators adopted in our experiments. Each of
these operators is based on a different encoding for the chromosome of a
candidate solution, which corresponds to a specific representation of an
$(n,k)$-combination. The reader is referred to Knuth~\cite{knuth-taocp} for
further information about the properties of these encodings. We emphasize that,
while the counter-based crossover operator is an adaptation of the one conceived
by Millan et al.~\cite{millan98}, to our knowledge the other two operators have
not been proposed before in the literature.

\subsection{Counter-Based Crossover}
\label{subsec:cnt-cross}
As discussed above, the \emph{binary vector coding} is the most obvious and
straightforward way to represent a $(n,k)$-combination: given a bitstring
$x = (x_1, \cdots, x_n)$ of length $n$, the positions of $x$ having value $1$
denote the $k$ selected objects out of a set of $n$, while the remaining $n-k$
zeros represent the unselected objects. As an example, consider the case where
$n=8$ and $k=4$. A $(8,4)$-combination can be represented by a balanced
bitstring of length $8$, such as: $x = (0,1,0,0,1,1,0,1)$.

Of course, binary vector coding is also the most natural chromosome
representation for GA. However, in order to evolve only individuals with a fixed
Hamming weight, one has to come up with a particular crossover operator. Perhaps
the simplest way to design such an operator is to randomly select bit-by-bit the
allele from the first or the second parent to be copied in the offspring (as in
uniform crossover), and use counters to keep track of the multiplicities of ones
and zero in the child. When one of the two counters reaches the prescribed
threshold (i.e. $k$ for the ones counter and $n-k$ for the zero counter), the
child is filled the complementary value.

To our knowledge, Millan et al.~\cite{millan98} were the first to propose a
crossover operator based on this idea to evolve nonlinear balanced Boolean
functions. We report in Algorithm~\ref{alg:cntcr} the pseudocode of a slightly
modified operator, which we used in our experiments.
\begin{algorithm}[t]
\floatname{algorithm}{Algorithm}
\caption{{\sc Counter-Cross}$(p_1, p_2, n, k)$}
\label{alg:cntcr}
\begin{algorithmic}
\STATE $s$ := $0$; $t$ := $0$; $c$ := $0^n$;
\FOR{$i$ := $1$ to $n$}
	\IF{($s$ = $k$)}
		\STATE $c[i]$ := $0$
	\ELSE
		\IF{($t$ = $n-k$)}
			\STATE $c[i]$ := $1$	
		\ELSE
			\STATE $c[i]$ := {\sc Random}($p_1[i]$, $p_2[i]$)
			\IF{($c[i]$ = $1$)}
				\STATE $s$ := $s+1$
			\ELSE
				\STATE $t$ := $t+1$
			\ENDIF
		\ENDIF
	\ENDIF
\ENDFOR
\STATE return $c$        
\end{algorithmic}
\end{algorithm}
Given two bitstrings $p_1, p_2 \in \F_2^n$ of length $n$ and Hamming weight $k$,
the procedure {\sc Counter-Cross} initializes the counters ($s$ for the number
of $1$s and $t$ for the number of $0$s) and sets to zero all the bits in the
string $c$, which will hold the chromosome of the child produced by the
crossover operation. Then, for $i \in \{1,\cdots,n\}$, the $i$-th bit of $c$ is
determined as follows. If the maximum number of ones (respectively, zeros)
allowed has already been reached, then $c[i]$ is set to $0$ (respectively,
$1$). In all other cases, $c[i]$ is chosen by randomly selecting with uniform
probability the $i$-th bit of $p_1$ or $p_2$, and the counters are updated
according to the drawn value. In this way, the child $c$ produced by {\sc
  Counter-Cross} is itself balanced.

\subsection{Zero Lengths Crossover}
\label{subsec:zerol-cross}
Given the bitstring $x = (x_1, \cdots, x_n)$ of a $(n,k)$-combination, the
\emph{zero lengths coding} of $x$ is the vector $r = (r_1, \cdots, r_{n-k+1})$
which lists the \emph{distances between consecutive ones} in $x$. In other
words, the values $r_i$ denote the lengths of the runs of zeros which separate
the ones in the binary vector coding, with the particular cases of $r_1$ and
$r_{n-k+1}$ which represent the number of zeros preceding the first $1$ and
following the last $1$ in $x$, respectively.

Clearly, in order to ensure that a given zero lengths coding vector
$r = (r_1, \cdots, r_{n-k+1})$ represents a valid $(n,k)$-combination, the
following relation must hold:
\begin{equation}
  \label{eq:rel-zerol}
\sum_{i=1}^{n-k+1} r_i = n-k \enspace .
\end{equation}
Following the example adopted in the previous two sections, the run length
coding of the bitstring $x$ is $r=(1,2,0,1,0)$. As we pointed out in
Equation~\eqref{eq:rel-zerol}, the zero lengths vector of a bitstring of length
$n$ and Hamming weight $k$ is valid if and only if the sum of the components in
the vector equals $n-k$. In a crossover operator based on the zero lengths
representation it is thus necessary to control the sum of the run lengths of
zeros in the offspring, while the components of the vector are copied from the
parents. The pseudocode for the crossover operator that we designed for this
specific coding is reported in Algorithm~\ref{alg:zerol}.
\begin{algorithm}[t]
\floatname{algorithm}{Algorithm}
\caption{{\sc Zero-Lengths-Cross}($p_1$, $p_2$, $n$, $k$)}
\label{alg:zerol}
\begin{algorithmic}
\STATE $sumz$ := $0$
\STATE $c$ := $0^{k+1}$

\FOR{$i$ := $1$ to $k$}
	\IF{($sumz = n-k$)}
		\STATE $c[i]$ := $0$
	\ELSE
                \STATE $cpar$ := {\sc Random}($p_1$, $p_2$)
                \IF{($sumz + cpar[i] = n-k$)}
                	\STATE $c[i]$ := $cpar[i]$
                        \STATE $sumz$ := $sumz + cpar[i]$
                \ELSE
                        \STATE $c[i]$ := $n - k - sumz$
                        \STATE $sumz$ := $n - k$
                \ENDIF
                
	\ENDIF
\ENDFOR
\STATE $c[k+1]$ := $n - k - sumz$
\STATE return $c$
\end{algorithmic}
\end{algorithm}
The operator takes as input the zero lengths vectors $p_1, p_2$ of two
bitstrings $x_1, x_2 \in \mathcal{B}_{n,k}$, their length $n$ and their Hamming
weight $k$. The first steps are devoted to the initialization of the zero length
vector of the child $c$ (filled with $k+1$ zeros) and the accumulator $sumz$
used to control the value of the sum of zeros in $c$. The FOR loop cycles over
the first $k$ positions of $c$. For each iteration $i$, an IF block initially
checks whether the sum of zeros in $c$ has already reached $n-k$, in which case
the value of $c[i]$ is set to zero. In the other case, a candidate parent $cpar$
is randomly selected between $p_1$ and $p_2$ with uniform probability. The next
IF block verifies whether the value at position $i$ of the selected parent
$cpar$ can be safely copied in the child without breaking the $n-k$ limit set by
Equation~\eqref{eq:rel-zerol}. If this is the case (i.e. $sumz + cpar[i]$ is at
most $n-k$), then $cpar[i]$ is copied in $c[i]$, and the accumulator $sumz$ is
updated by adding to it $cpar[i]$. Otherwise, the value in $c[i]$ is set to the
remaining number of zeros that can be put in the child, which is
$n-k-sumz$. Equivalently, this means that we are copying in $c[i]$ just enough
zeros from $cpar[i]$ to reach the threshold $n-k$, without violating it. Then,
the accumulator $sumz$ is directly set to $n-k$, since no other zeros can be put
in the child. After the FOR loop, the value of the last component in $c$ is
determined by simply subtracting from $n-k$ the sum of zeros obtained up to that
point. Thus, if $sumz$ reached $n-k$ in the FOR loop, the last component will be
set to zero, otherwise it will contain the number of zeros necessary to ``pad''
the bitstring encoded by $c$ after its last $1$.

\subsection{Map of Ones Crossover}
\label{subsec:map1-cross}
Suppose that $x = (x_1, \cdots, x_n)$ is the binary vector representation of a
$(n,k)$-combination, and denote by $supp(x)$ its support. The \emph{map of ones}
of $x$ is the $k$-dimensional vector $q=(q_1, \cdots, q_k)$ where
$q_i \in supp(x)$ for all $i \in \{1,\cdots,k\}$, and such that $q_i\neq q_j$
for all indices $i\neq j$. In other words, the map of ones of $x$ corresponds to
its support in vector form.

Thus, the map of ones representation lists the nonzero coordinates in the binary
coding of a $(n,k)$-combination. Following the example of the previous section,
the map of ones corresponding to the binary string $x$ representing a
$(8,4)$-combination is $q=(2,5,6,8)$, where the positions of the ones are listed
in increasing order. Strictly speaking, the order of the positions is
irrelevant, since they always yield the same binary representation.

One can notice that the only constraint in the map of ones is that there cannot
be duplicate positions in the vector. Thus, given two bitstrings of length $n$
and weight $k$ represented by their maps of ones, the crossover operator must be
aware of the common positions between them, in order to avoid
duplications. Algorithm~\ref{alg:map1cr} reports the pseudocode for our
crossover operator.
\begin{algorithm}[t]
\floatname{algorithm}{Algorithm}
\caption{{\sc Map-1-Cross}($p_1$, $p_2$, $k$)}
\label{alg:map1cr}
\begin{algorithmic}
\STATE $c$ := $0^{k}$
\STATE $comm\_list$ = {\sc Find-Common-Pos}($p_1$, $p_2$)

\FOR{$i$ := $1$ to $k$}
	\STATE $cpar$ := {\sc Random}($p_1$, $p_2$)
	\STATE $cpos$ := {\sc Rand-Pos}($cpar$)
	\STATE $c[i]$ := $cpar[cpos]$
	\STATE {\sc Remove}($cpar$, $cpar[cpos]$)
	\IF{ ({\sc Contains}($comm\_list$, $cpar[cpos]$)) }
		\IF{ ($cpar$ = $p_1$) }
			\STATE {\sc Remove}($p_2$, $cpar[cpos]$)
		\ELSE
			\STATE {\sc Remove}($p_1$, $cpar[cpos]$)
		\ENDIF
	\ENDIF
\ENDFOR
\STATE return $c$        
\end{algorithmic}
\end{algorithm}
Let us suppose that we have two bitstrings $x_1,x_2 \in \mathcal{B}_{n,k}$ of
length $n$ and Hamming weight $k$, represented respectively by the maps of ones
$p_1$ and $p_2$ of length $k$. The procedure {\sc Map-1-Cross} begins by
initializing the map of ones of the child $c$ and by finding the positions which
$p_1$ and $p_2$ have in common. The latter operation is performed by the
subroutine {\sc Find-Com-Pos}, which returns the vector
$comm\_list$. Successively, for all $i \in \{1,\cdots,k\}$, the value $c[i]$ is
computed as follows. One of the two parents is randomly chosen by calling the
procedure {\sc Random} on $p_1$ and $p_2$. Then, a random index $cpos$ is
selected from the candidate parent $cpar$, and the value of $c[i]$ is set equal
to $cpar[cpos]$. In other words, the child $c$ inherits from the parent $cpar$
the position of the $1$ specified by the value $cpar[cpos]$. Finally, in order
to avoid that the same position is selected in the next iterations, the value
$cpar[cpos]$ is removed from the candidate parent by using the {\sc Remove}
procedure. The value is also removed from the unselected parent if it is
contained in $comm\_list$.

\subsection{Ordering Bias and Positions Shuffling}
\label{subsec:ord-bias}
It can be noticed that all the balanced crossover operators that we defined in
the previous sections build the child individual from left to right, i.e. by
copying the genes from the parents in increasing order of position. A natural
question arising from this observation is whether this particular ordering
introduces any bias towards a particular subset of feasible
offspring. Intuitively, this does not seem to be the case for the map-of-ones
crossover, since as we remarked in Section~\ref{subsec:map1-cross} the phenotype
bitstring depends only on the specific values contained in the map of ones
encoding, and not on their ordering. On the other hand, for the counter-based
and zero-length crossover operators the situation looks different. In fact,
after the threshold value has been reached (be it the number of $0$ or $1$, or
the sum of zero lengths), the remaining loci of the child are set
deterministically. Hence, it would be reasonable to assume that the
counter-based and the zero-length crossover operators induces a bias in the
offspring. We remark that, although this ``left-to-right'' approach is prevalent
in the relevant literature since Millan et al's counter-based
crossover~\cite{millan98}, to the best of our knowledge no one investigated the
impact of this design choice on the GA performance.

For this reason, we also considered a ``shuffling'' version of each balanced
crossover operator which randomly mixes the order of the positions to be copied
from the parents to the offspring, to assess if there are significant
differences in performances with the ``left-to-right'' approach. From the
pseudocode point of view, the shuffling version of each balanced operator is
practically identical to its ``left-to-right'' counterpart, except that an
additional array $pos$ of length $n$ representing a random permutation of the
genotype positions is passed as an input parameter to the operator. Moreover,
all occurrences of $i$ used to index the positions of the child $c$ or the
parents $p_1, p_2$ are replaced by $pos[i]$.

\section{Optimization Problems}
\label{sec:problems}
We now give the formal statement of the three combinatorial optimization
problems that we addressed in our statistical comparison of balanced crossover
operators.

\subsection{Nonlinear Balanced Boolean Functions}
\label{subsec:nlin-bal-bent}
A \emph{Boolean function} of $n \in \N$ variables is a map
$f: \F_2^n \rightarrow \F_2$. The common way for representing a Boolean function
$f$ is by means of its \emph{truth table} $\Omega_f$, which is basically a
binary vector of length $2^n$ that specifies for each input vector
$x \in \F_2^n$ the output value of $f(x)$, in lexicographic order. A Boolean
function is called \emph{balanced} if its \emph{truth table} $\Omega_f$ is
composed of an equal number of ones and zeros, i.e. it represents a
$(2^n,2^{n-1})$-combination.

Another representation of Boolean functions $f: \F_2^n \rightarrow \F_2$ used in
cryptography is the \emph{Walsh transform}, which is the function
$W_f: \F_2^n \rightarrow \Z$ defined for all $\omega \in \F_2^n$ as:
\begin{equation}
  \label{eq:wt}
  W_f(\omega) = \sum_{x \in \F_2^n} (-1)^{f(x)}\cdot (-1)^{\omega\cdot x}
  \enspace ,
\end{equation}
where
$\omega\cdot x = \omega_1x_1 \oplus \omega_2x_2 \oplus \cdots \oplus
\omega_nx_n$ is the \emph{scalar product} modulo $2$ between the vectors
$\omega, x \in \F_2^n$. The \emph{spectral radius} $W_{max}(f)$ of a Boolean
function $f$ is defined as the maximum absolute value of its Walsh transform,
i.e.  $W_{max}(f) = \max_{\omega \in \F_2^n}\{|W_f(\omega)\}$.

The \emph{nonlinearity} of a Boolean function $f:\F_2^n \rightarrow \F_2$ is
defined as the minimum Hamming distance of its truth table $\Omega_f$ from the
set of truth tables of all linear functions, i.e. those functions whose
algebraic expressions contain only XOR. This can be computed through the
following formula based on the Walsh transform:
\begin{equation}
  \label{eq:nl-bf}
  Nl(f) = 2^{n-1} - \frac{1}{2} \cdot W_{max}(f) \enspace .
\end{equation}

In cryptography, Boolean functions which are both balanced and have high
nonlinearity play a fundamental role in the design of stream and block
ciphers~\cite{carlet-bool}. Since the set of all Boolean functions is composed
of $2^{2^n}$ elements, which is not exhaustively searchable for $n>5$,
evolutionary algorithms such as GA represent a possible method for finding
highly nonlinear balanced Boolean functions in a reasonable amount of time. We
formally state the combinatorial optimization problem as follows:
\begin{problem}
  \label{prob:nlin-bf}
  Let $n \in \N$. Find a Boolean function $f: \F_2^n \rightarrow \F_2$ of $n$
  variables such that $f$ is balanced and has maximum nonlinearity.
\end{problem}
In particular, given the truth table bitstring $\Omega_f \in \F_2^{2^n}$ of a
Boolean function $f: \F_2^n \rightarrow \F_2$ of $n$ variables, in our
experiments the fitness of $f$ is computed with the following function:
\begin{equation}
  \label{eq:fit1}
\fit_1(f) = Nl(f) - |2^{n-1} - w_H(\Omega_f)| \enspace ,
\end{equation}
where $|2^{n-1} - w_H(\Omega_f)|$ is the \emph{unbalancedness penalty factor}
which punishes the deviation of $f$ from being a balanced function.  The
objective of our GA, in particular, is to \emph{maximize} $\fit_1(f)$. Of
course, when using balanced crossover operators the penalty factor is not
necessary, since the candidate solutions generated by GA are always balanced
functions.

\subsection{Bent Functions}
\label{subsec:bent}
From Equation~\eqref{eq:nl-bf}, one can see that the lower the spectral radius
is, the higher the nonlinearity of a Boolean function will be. Due to
\emph{Parseval's relation}~\cite{carlet-bool}, the minimum spectral radius is
achieved when the Walsh spectrum is uniformly divided among all $2^{n}$
vectors. This means that the Walsh coefficients must all have the same absolute
value $2^{\frac{n}{2}}$, thus giving the following upper bound on nonlinearity:
\begin{equation}\label{eq:cov-bound}
Nl(f) \le 2^{n-1} - 2^{\frac{n}{2} - 1} \enspace .
\end{equation}
\noindent
Clearly, equality in~\eqref{eq:cov-bound} can occur only if $n$ is even, since
the Walsh coefficients of a Boolean function must be integer numbers. The
Boolean functions achieving this bound are called \emph{bent}, and they have
several applications in cryptography and coding theory~\cite{carlet-bool}.

A nice feature of the Walsh transform is that the Walsh coefficient $W_f(0)$
(where $0$ denotes the null vector) is related to the Hamming weight of the
truth table $\Omega_f$ as follows:
\begin{equation}
  \label{eq:hwt-wt}
  w_H(\Omega_f) = 2^{n-1} - \frac{1}{2} \cdot W_f(0) \enspace .
\end{equation}
Since all Walsh coefficients of a bent function must be equal to
$\pm 2^{\frac{n}{2}}$, this means that the Hamming weight of bent functions is
either $2^{n-1} - 2^{\frac{n}{2}-1}$ or $2^{n-1} + 2^{\frac{n}{2}-1}$. Without
loss of generality, one can narrow the attention only to the weight
$2^{n-1} - 2^{\frac{n}{2}-1}$, since the others are obtained by simply
complementing the corresponding truth tables. Hence, one can cast the search of
bent functions as an optimization problem over the set of bitstrings of length
$2^n$ and weight $k = 2^{n-1} - 2^{\frac{n}{2}-1}$, which makes it amenable to
GA with balanced crossover operators. For this reason, we adopted it as our
second optimization problem for our investigation:
\begin{problem}
  \label{prob:bent}
  Let $n \in \N$ be an even number. Find a Boolean function
  $f: \F_2^n \rightarrow \F_2$ of $n$ variables such that
  $Nl(f) = 2^{n-1} - 2^{\frac{n}{2} - 1}$.
\end{problem}

Since bent functions reach the highest possible value of nonlinearity, we
defined a fitness function analogous to $fit_1$. Given
$f: \F_2^n \rightarrow \F_2$, we defined the fitness function over $f$ for
Problem~\ref{prob:bent} as follows:
\begin{equation}
  \label{eq:fit2}
  \fit_2(f) = Nl(f) - |2^{n-1} - 2^{\frac{n}{2}-1} - w_H(\Omega_f)| \enspace ,
\end{equation}
where the unbalancedness penalty factor this time is defined as
$|2^{n-1} - 2^{\frac{n}{2}-1} - w_H(\Omega_f)|$. The optimization objective is
again to \emph{maximize} Equation~\eqref{eq:fit2}, since having $\fit_2(f)$
equal to the covering bound for $n$ corresponds to the case where the
nonlinearity is maximal and the deviation from the prescribed Hamming weight
$2^{n-1} - 2^{\frac{n}{2}-1}$ is zero. As in the case of $\fit_1$, when using
balanced crossover operators the unbalancedness penalty factor is not necessary.

\subsection{Binary Orthogonal Arrays}
\label{subsec:oa}
\emph{Orthogonal Arrays} (OA) are rectangular matrices whose submatrices satisfy
a specific balancedness constraint on their rows. OA find several applications
in statistics, combinatorial designs theory and cryptography~\cite{hedayat}. In
what follows, we will focus on \emph{binary} OA, meaning that the matrices are
Boolean. We formally define a binary OA as follows:
\begin{definition}
  \label{def:bin-oa}
  Let $N, k, t, \lambda \in \N$ with $0 \le t \le k$. A $N \times k$ binary
  matrix $A$ is called a binary \emph{orthogonal array} (OA) with $k$ columns,
  strength $t$ and index $\lambda$ (for short, an $OA(N,k,t,\lambda)$) if in
  each submatrix of $N$ rows and $t$ columns each binary $t$-tuple occurs
  exactly $\lambda$ times.
\end{definition}
Notice that the parameter $\lambda$ of an OA is related to its strength $t$ and
number of rows $N$ by the relation $\lambda = \frac{N}{2^t}$.

For our third optimization problem, formally defined below, we are interested in
binary OA whose columns are truth tables of Boolean functions:
\begin{problem}
  \label{prob:bin-oa}
  Let $n, k, t \in \N$. Find $k$ Boolean functions $f_1,\cdots,f_k:\F_2^n
  \rightarrow \F_2$ of $n$ variables such that the matrix
  \begin{equation}
    A = [\Omega_{f_1}^{\top}, \Omega_{f_2}^{\top}, \cdots, \Omega_{f_k}^{\top}]
  \end{equation}
  is an $OA(2^n, k, t, \lambda)$, with $\lambda = 2^{n-t}$.
\end{problem}
Hence, the goal of Problem~\ref{prob:bin-oa} is to find $k$
$n$-variables Boolean functions such that the bitstrings of their truth tables
are the columns of a binary OA with $N=2^n$ rows and strength $t$.

A useful property for this problem is that any binary OA of strength $t$ is also
an OA of strength $t'<t$, for all $t' \in \{1,\cdots,t-1\}$. Taking $t'=1$, this
implies that each column of a binary OA must be a balanced bitstring of length
$N$. Consequently, one can use a GA with balanced crossover operators to evolve
candidate binary OA as a set of $k$ balanced bitstrings. New solutions are bred
by applying balanced crossover and mutation independently on the single
bitstrings, thus maintaining the balancedness constraint on the single columns
of the array. This is the optimization approach that was adopted by Mariot et
al.~\cite{mariot18}, from which we took the fitness function for our
experiments. In particular, the fitness function stands on the idea of counting
the \emph{repeated tuples} in the submatrices of an array.

Given a $N\times k$ binary matrix $A$, let $I$ be a subset of $t$ indices, and
let $A_I$ denote the $N \times t$ submatrix obtained by considering only the
columns of $A$ specified by the indices of $I$. For all binary $t$-tuples
$x \in \F_2^t$, let $A_I[x]$ denote the number of occurrences of $x$ in $A_I$,
and let $\delta(A_I, x)$ be the $\lambda$-\emph{deviation} of $x$ defined as
$\delta(A_I, x) = | \lambda - A_I[x] |$. Then, the \emph{Euclidean deviation} of
$A_I$ is defined as:
\begin{equation}
\label{eq:dev-eu}
\Delta(A_I)_2 = \sqrt{\left ( \sum_{x \in \F_2^t} \left | \lambda - A_I[x] \right
  |^2 \right )} \enspace .
\end{equation}
The fitness function for Problem~\ref{prob:bin-oa} is then defined for all
$2^n\times k$ binary matrix $A$ formed by $k$ $n$-variables Boolean functions as
follows:
\begin{equation}
\label{eq:fit3}
\fit_3(A) = \sum_{I \subseteq [k]: |I| = t} \Delta(A_I)_2 + UNB(A) \enspace ,
\end{equation} 
where the unbalancedness penalty factor $UNB(A)$ is defined as the sum of the
unbalancedness of all Boolean functions $f_1,\cdots,f_k$, that is,
$UNB(A) = \sum_{i=1}^k |2^{n-1} - w_H(\Omega_{f_i})|$. As for the other two
optimization problems, when using GA with balanced crossover operators this
penalty factor can be dropped from the fitness function. The optimization
objective is to minimize $\fit_3$, since any binary matrix such that
$\fit_3(A) = 0$ corresponds to a binary $OA(2^n, k, t, \lambda)$.

\section{Experimental Setting}
\label{sec:exp-set}

\label{sec:exp-set}
In this section, we describe the details of the genetic algorithm used to test
the three crossover operators presented in Section~\ref{sec:crossover}, the
parameters used to set up the experiments over the three optimization problem
defined in Section~\ref{sec:problems}, and the experimental hypotheses to be
tested\footnote{The source code of our GA and the experimental results presented
  in the next section are publicly available at
  \url{https://github.com/rymoah/BalancedCrossoverGA}}.

\subsection{Genetic Algorithm Details}
\label{subsec:ga-det}
The genetic algorithm adopted in this work is a \emph{steady state} GA where a
single pair of parents is drawn from the current population at each
iteration. For selection, we employed a \emph{deterministic tournament} operator
where the best two out of $t$ randomly sampled individuals are selected for
crossover.

The four crossover operators considered in our investigation are classic
one-point crossover and the three balanced crossover described in
Section~\ref{sec:crossover}, namely counter-based, map of ones and zero-lengths
crossover. Our GA generates a single child for each selected pair of parents,
independently of the underlying crossover operator. In particular, since
one-point crossover generates two children by design, our GA randomly selects
only one of them with uniform probability, which is then subjected to mutation.

The mutation operator depends on the type of crossover: when one-point crossover
is used, a classic bit-flip mutation operator is applied on the generated
child. On the other hand, with balanced crossover operators a simple
\emph{swap-based} mutation operator is used~\cite{mariot18}. In particular, this
mutation operator swaps with probability $p_m$ a pair of distinct values in the
child, in order to maintain its Hamming weight. Notice that swap mutation always
operates on the bitstring representation. Hence, even when zero-length or
map-of-ones crossover are employed, the child is first mapped to the
corresponding balanced bitstring, and then swap mutation is applied to
it. Further, remark that in the OA problem the mutation operator (be it bit-flip
or swap mutation) is applied separately on each column composing the child, as
done in~\cite{mariot18}.

Once the child has been mutated, the GA evaluates the relevant fitness function
on it. In particular, depending on the optimization problem considered, if
one-point crossover is used then the full form of the fitness functions $fit_1$,
$fit_2$ or $fit_3$ respectively described in
Equations~\eqref{eq:fit1},~\eqref{eq:fit2} and~\eqref{eq:fit3} is used. On the
opposite, when one of the three balanced crossover operators is employed, the
unbalancedness penalty factor is dropped from the computation of the fitness
functions. Similarly to the choice of the mutation operator, the creation of the
initial population depends on the adopted crossover operator. For one-point
crossover, the population is initialized at random, without controlling the
Hamming weights of the generated bitstrings. Contrarily, for balanced crossover
operators each chromosome is created by generating at random each bit in the
bitstring, and using a counter to keep track of the number of ones. When the
prescribed Hamming weight has been reached, the chromosome is filled with zeros
in the remaining positions.

Our GA uses an elitist strategy with random replacement: in particular, if the
child has a better fitness value than any of its two parents, then an individual
is drawn at random from the population to be replaced by the child. To guarantee
elitism, if the child has a better fitness value of its two parents but not of
the best individual in the whole population, then the latter is excluded from
the replacement. Further, since we are more interested in comparing the
performances of the crossover operators than in generating optimal solutions for
the considered problems, the GA terminates after it performs a specified number
of fitness evaluations.

\subsection{Experimental Parameters}
\label{subsec:exp-par}
Table~\ref{tab:pb} reports the problem instances tested in our experiments for
each of the three considered optimization problems.
\begin{table}[t]
\centering
\caption{Problem instances and relative search space sizes}
\begin{tabular}{p{1cm}p{2cm}p{3cm}p{3cm}}
\hline\noalign{\smallskip}
Problem & Instance & UNB Size & BAL Size \\
\noalign{\smallskip}\hline
\noalign{\smallskip}
  \multirow{3}{*}{{\sc Bal-NL}}    & $n=6$    & $2^{2^6}\approx 1.8\cdot 10^{19}$  & $\binom{64}{32}\approx 1.8\cdot 10^{18}$   \\
  \noalign{\smallskip}
                                   & $n=7$    & $2^{2^7}\approx 3.4\cdot 10^{38}$ & $\binom{128}{64}\approx 2.4\cdot 10^{37}$  \\
  \noalign{\smallskip}
                                   & $n=8$    & $2^{2^8}\approx 1.1\cdot 10^{77}$ & $\binom{256}{128}\approx 5.7\cdot 10^{75}$ \\
  \noalign{\smallskip}\hline
  \multirow{3}{*}{{\sc Bent}}      & $n=6$     & $2^{2^6}\approx 1.8\cdot 10^{19}$    & $\binom{64}{28}\approx 1.1\cdot 10^{18}$   \\
  \noalign{\smallskip}
                                   & $n=8$     & $2^{2^8}\approx 1.1\cdot 10^{77}$   & $\binom{256}{120}\approx 3.5\cdot 10^{75}$  \\
  \noalign{\smallskip}
                                   & $n=10$    & $2^{2^{10}}\approx 1.8\cdot 10^{308}$ & $\binom{1024}{496}\approx~2.7\cdot~10^{306}$ \\
  \noalign{\smallskip}\hline
  \multirow{3}{*}{{\sc Bin-OA}}    & $OA(16, 8, 3, 2)$  & $\binom{2^{16}}{8}\approx 8.4\cdot 10^{33}$  & $\binom{\binom{16}{8}}{8}\approx 1.8\cdot 10^{28}$  \\
  \noalign{\smallskip}
                                   & $OA(16, 8, 2, 4)$  & $\binom{2^{16}}{8}\approx 8.4\cdot 10^{33}$  & $\binom{\binom{16}{8}}{8}\approx 1.8\cdot 10^{28}$  \\
  \noalign{\smallskip}
                                   & $OA(16, 15, 2, 4)$ & $\binom{2^{16}}{15}\approx 1.3\cdot 10^{60}$ & $\binom{\binom{16}{8}}{15}\approx 3.3\cdot 10^{49}$ \\
  
\noalign{\smallskip}\hline
\noalign{\smallskip}
\end{tabular}
\label{tab:pb}
\end{table}
The three problems in the table are respectively identified in the first column
by the names {\sc Bal-NL} for highly nonlinear balanced Boolean functions, {\sc
  Bent} for bent functions and {\sc Bin-OA} for binary orthogonal arrays. The
second column reports the problem instances considered for each problem, which
are characterized by the number of variables $n$ for the problems {\sc Bal-NL}
and {\sc Bent} and by the set of parameters $OA(2^n, k, t, \lambda)$ for the
{\sc Bin-OA} problem. The third and fourth columns report, for each problem
instance, the size of the corresponding search space respectively when one-point
crossover is used (UNB Size) and when balanced crossover operators are adopted
(BAL size). The details for the computation of the search space sizes can be
found in~\cite{mariot18}. It can be remarked from Table~\ref{tab:pb} that for
the two Boolean functions problems ({\sc Bal-NL} and {\sc Bent}) the use of
balanced crossover operators yields a reduction of the search space between one
and two orders of magnitude. On the other hand, for the {\sc Bin-OA} problem the
reduction is much more significant.

For each problem instance, we ran our steady state GA with each of the four
crossover operators for $R=50$ experimental runs. Moreover, for each balanced
operator we performed a separate set of $R=50$ runs both for the
``left-to-right'' and the ``shuffling'' versions. Hence, we tested $7$ crossover
operators for a total of $7 \cdot 3 \cdot 50 = 1050$ experimental runs for each
of the three optimization problems. Each GA experiment used a population size of
$P=50$ individuals, tournament size $t=3$ and stopped after $fit=500000$ fitness
evaluations. Additionally, we employed a mutation probability of $0.7$ for the
{\sc Bal-NL} and the {\sc Bent} problems and of $0.2$ for the {\sc Bin-OA}
problem. We chose these particular parameter values in order to compare our
balanced Boolean functions and bent functions results with those reported
in~\cite{picek17}, and our OA results with those of~\cite{mariot18}.

The research hypotheses that we tested with our experiments are as follows:
\begin{itemize}
\item $H_1$: One-point crossover has a worse performance than any of the three
  balanced crossover operators.
\item $H_2$: The shuffling versions of the counter-based and zero-length
  crossover have a better performance than their ``left-to-right'' counterparts.
\item $H_3$: There is no statistical significant difference between the
  shuffling and the ``left-to-right'' versions of the map-of-ones crossover.
\end{itemize}

In order to settle the research hypotheses above when observation of the results
plots was not sufficient, we employed the \emph{Mann-Whitney-Wilcoxon test}. The
alternative hypothesis adopted was that the two distributions were not
equal. More precisely, that the probability of a sample $a$ from the first
distribution exceeding a sample $b$ from the second distribution is different
from the probability of $b$ exceeding $a$. The significance value $\alpha$ for
the statistical tests was set to $0.01$.

\section{Results}
\label{sec:res}
\begin{figure}
  \centering
  \includegraphics[scale=0.5]{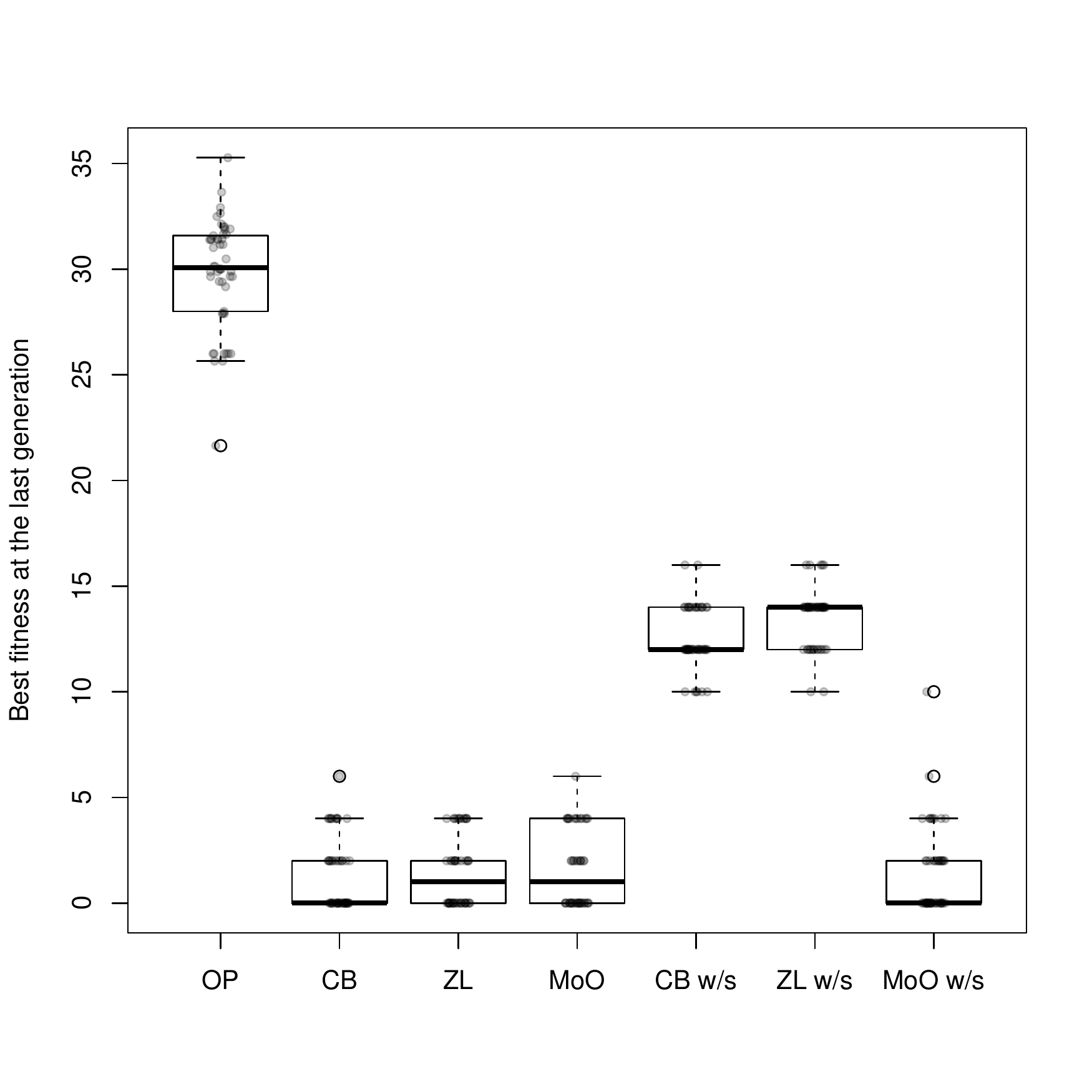}
  \caption{Results of the OA problem, $(16,8,2,4)$ instance.}
  \label{fig:oa-ist1}
  \includegraphics[scale=0.5]{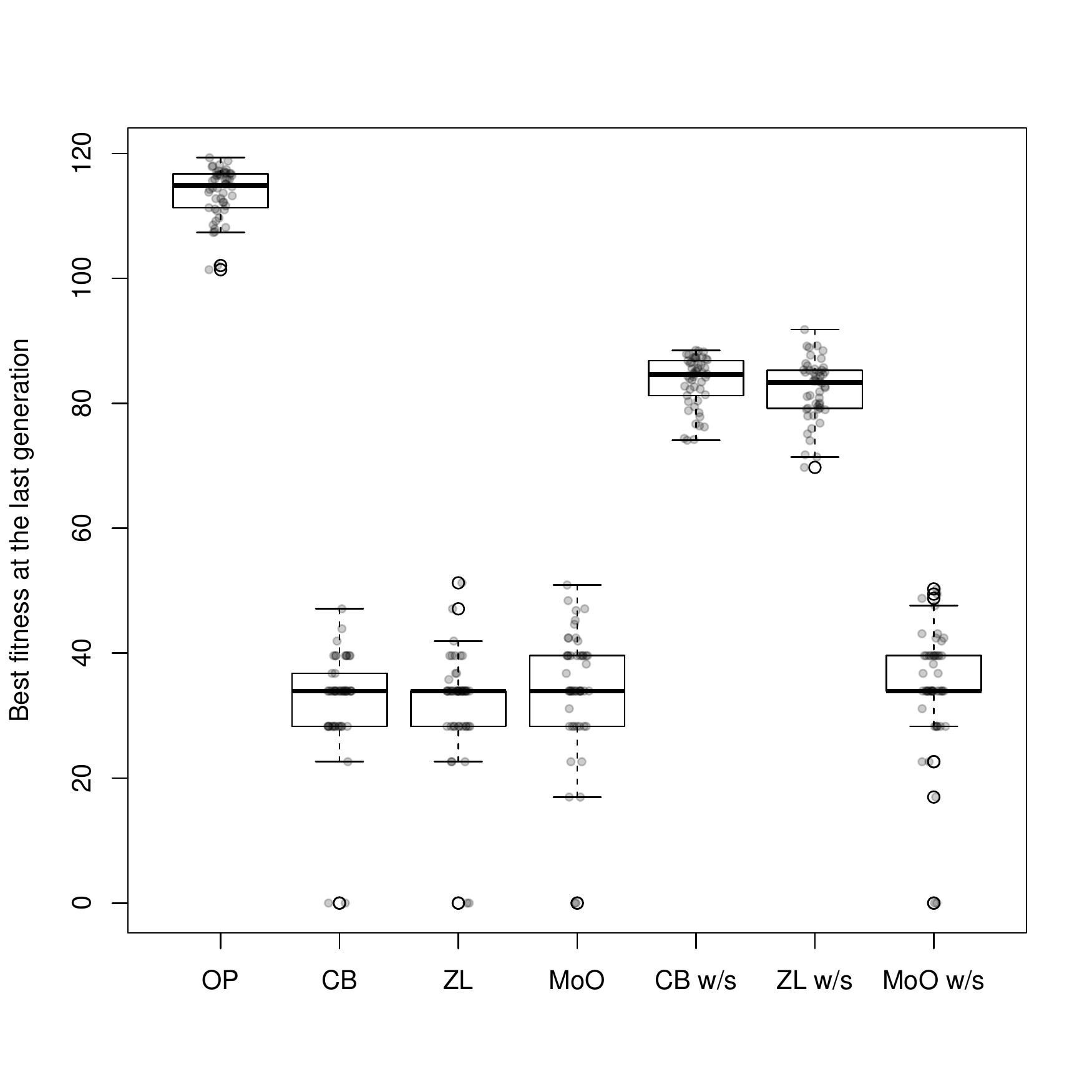}
  \caption{Results of the OA problem, $(16,8,3,2)$ instance.}
  \label{fig:oa-ist2}
\end{figure}

\begin{figure}
  \centering
  \includegraphics[scale=0.5]{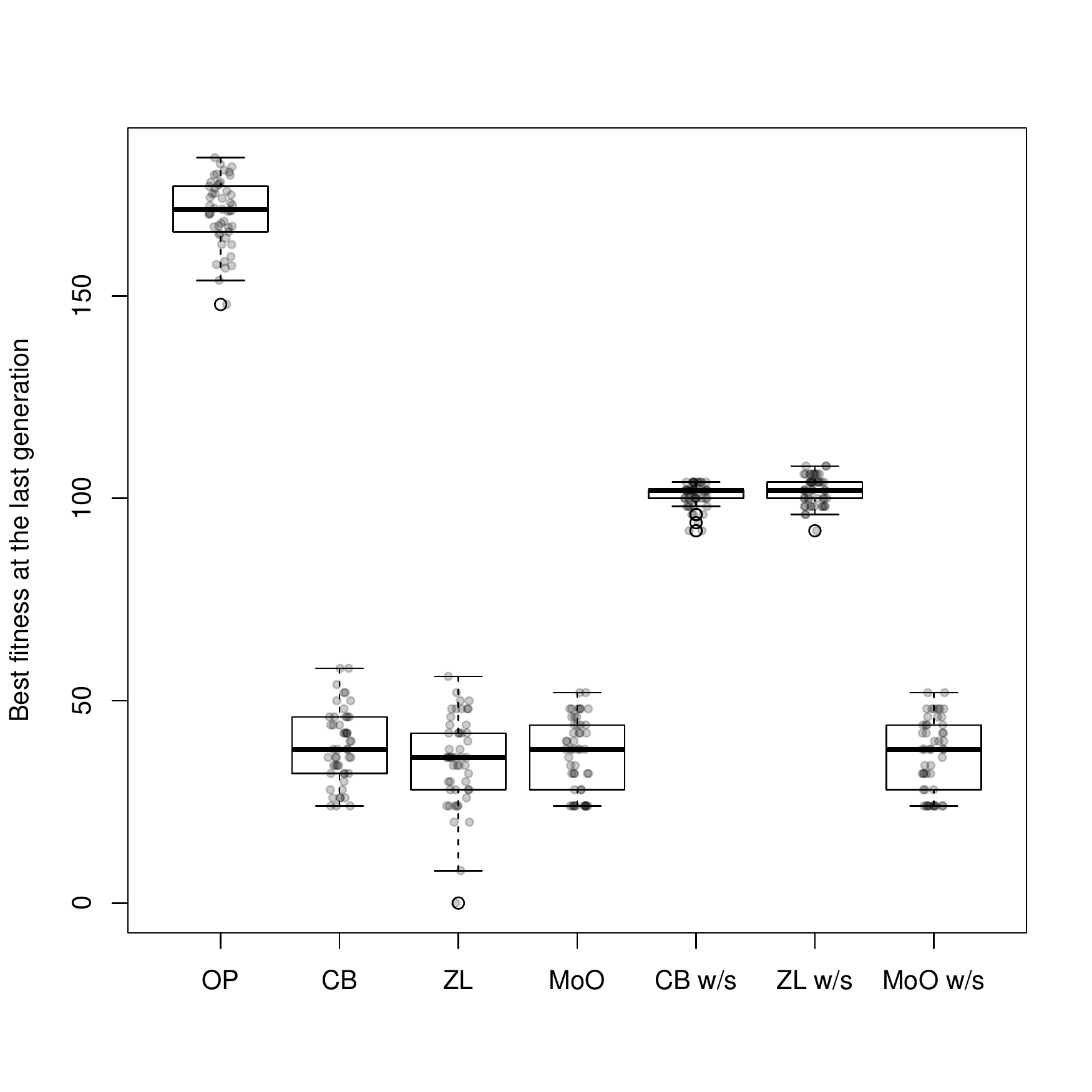}
  \caption{Results of the OA problem, $(16,15,2,4)$ instance.}
  \label{fig:oa-ist3}
\end{figure}

The results of the experiments are summarized in
Figures~\ref{fig:bf-nvar6}--~\ref{fig:oa-ist3}. Each figure represents one of
the three problem instances of the three considered optimization problems. The
boxplots show the median, minimum, maximum, first and third quartiles (excluding
outliers) of the fitness values of the best individuals reached at the end of
the experimental runs with a particular crossover operator. In all figures, the
acronyms OP, CP, ZL and MoO respectively stand for one-point, counter-based,
zero-length and map-of-ones crossover, with the ``w/s'' suffix indicating the
shuffling versions of the balanced operators.

\subsection{Balanced Boolean Functions}
\label{subsec:res-bf}
\begin{figure}
  \centering
  \includegraphics[scale=0.5]{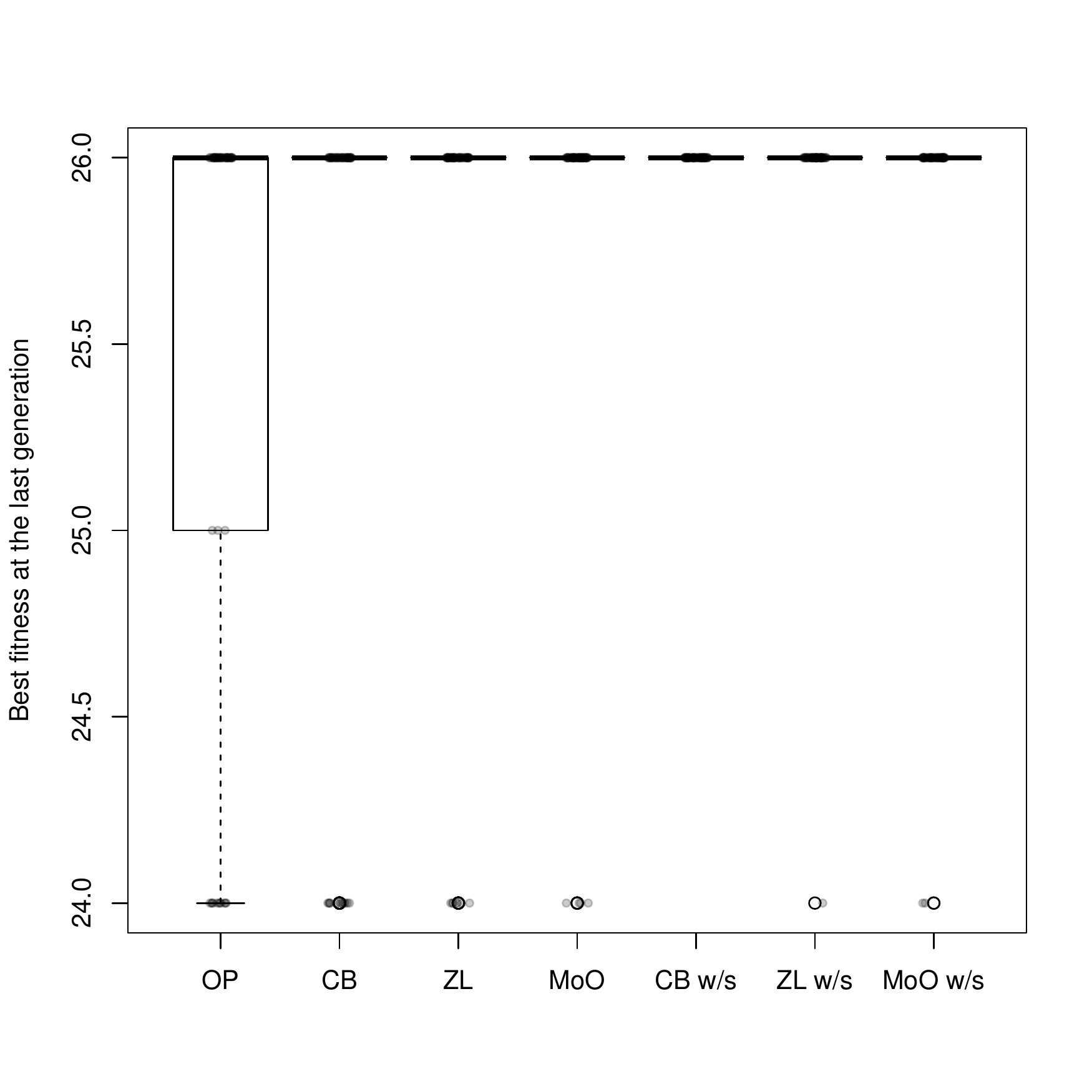}
  \caption{Results of the Balanced Boolean functions problem, $n=6$ instance.}
  \label{fig:bf-nvar6}
  \includegraphics[scale=0.5]{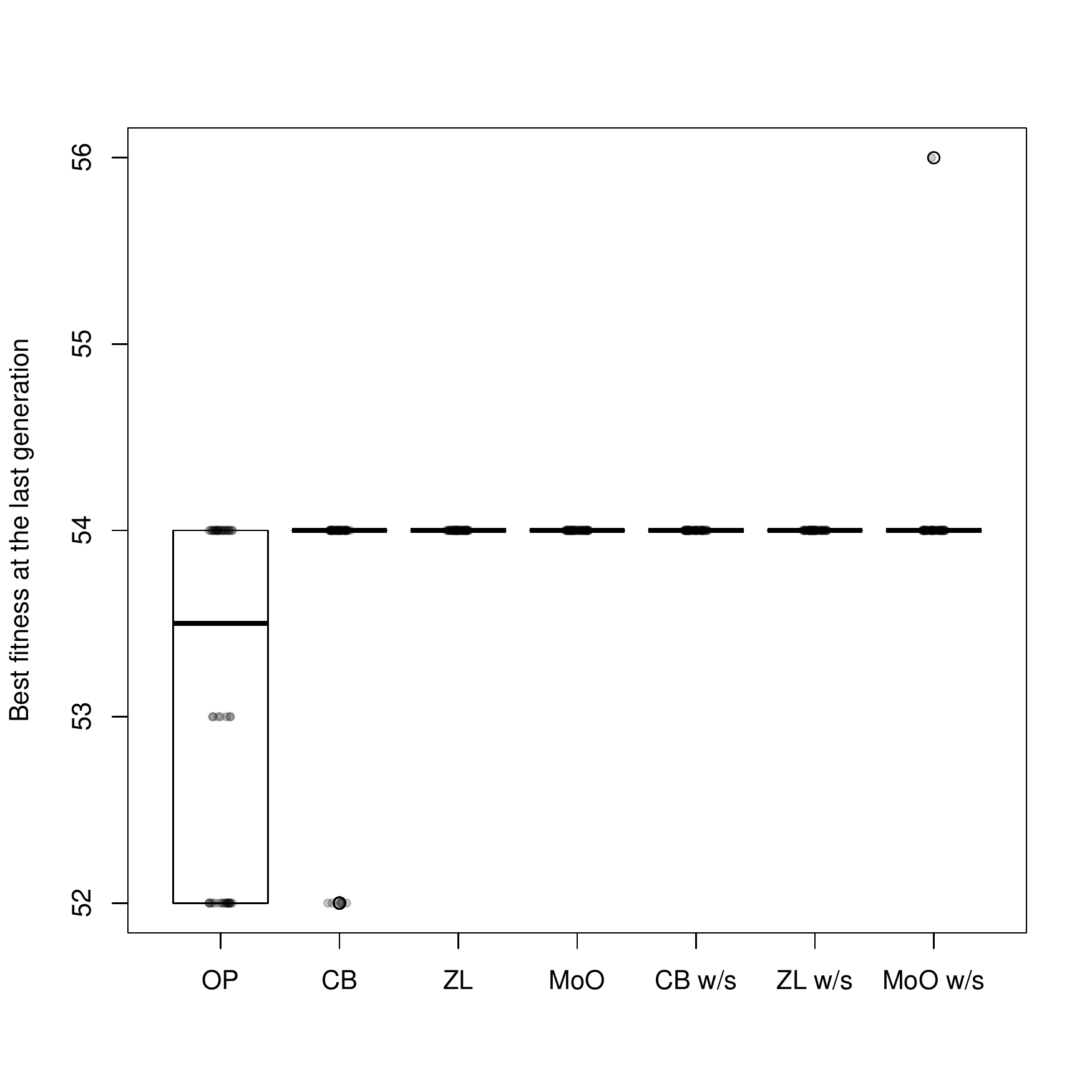}
  \caption{Results of the Balanced Boolean functions problem, $n=7$ instance.}
  \label{fig:bf-nvar7}
\end{figure}

\begin{figure}
  \centering
  \includegraphics[scale=0.5]{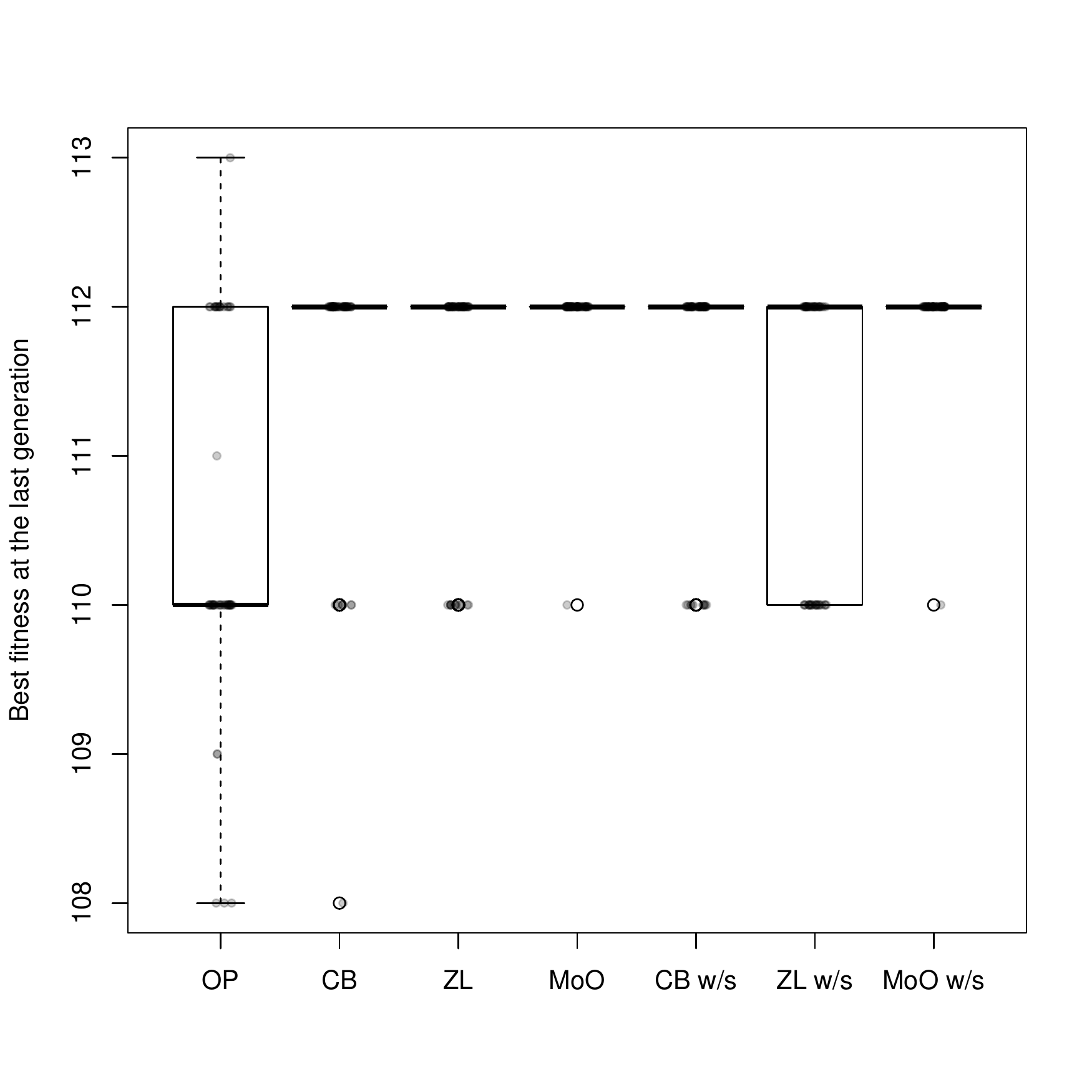}
  \caption{Results of the Balanced Boolean functions problem, $n=8$ instance.}
  \label{fig:bf-nvar8}
\end{figure}

Figures~\ref{fig:bf-nvar6}--~\ref{fig:bf-nvar8} depict the results obtained for
the balanced Boolean functions problem.


From the results for $n=6$, it is possible to observe that all crossover
operators have the same median fitness of $26$, which corresponds to the maximum
nonlinearity achievable by balanced functions of $6$ variables. Although the
results obtained by one-point crossover seem to be more dispersed towards lower
values of nonlinearity, the statistical tests show no significant differences
between its distribution and those of the ``left-to-right'' versions of the
balanced operators. On the other hand, one-point crossover performs worse than
the shuffling versions of all balanced operators ($p$-values of
$6.30 \cdot 10^{-5}$, $0.0003$, and $0.00013$ respectively for the comparisons
OP-CB w/s, OP-ZL w/s and OP-MoO w/s). On the other hand, for the $n=7$ instance
one-point crossover performs worse than all other balanced crossover operators,
both shuffling and left-to-right versions. Beside being confirmed by the
statistical tests, this can also be remarked by the boxplots in
Figure~\ref{fig:bf-nvar7}, where the median fitness scored by OP is lower than
the medians of the balanced operators. The situation is about the same for $n=8$
variables, where OP performs worse than all other balanced operators except when
comparing it with the shuffling version of the zero-length crossover, since the
statistical tests did not yield any significant difference between their
performances. Remark that, although one-point crossover produced a maximum best
fitness of $113$ which is higher than the one scored by the balanced operators,
this does not correspond to a balanced solution (since the nonlinearity of
balanced functions must be even).

Regarding the comparison between the left-to-right and shuffling versions of the
balanced crossover operators over this problem, we detected a statistically
significant difference only for the counter-based crossover over the $n=6$ and
$n=7$ instances, with the shuffling version performing better ($p$-values of
$0.0002$ and $0.0065$ respectively). In particular, the counter-based crossover
with shuffle is the only operator reaching a $100\%$ success rate for the $n=6$
instance, since all best individuals achieved the maximum fitness value of
$26$. However, when considering all three problem instances, the map-of-ones
crossover seems to be the most robust operator, since from the boxplots it has
the smallest dispersion. Moreover, the shuffling version of the map-of-one
operator achieved the maximum nonlinearity value of $56$ for balanced Boolean
functions of $n=7$ variables.

\subsection{Bent Functions}
\begin{figure}
  \centering
  \includegraphics[scale=0.5]{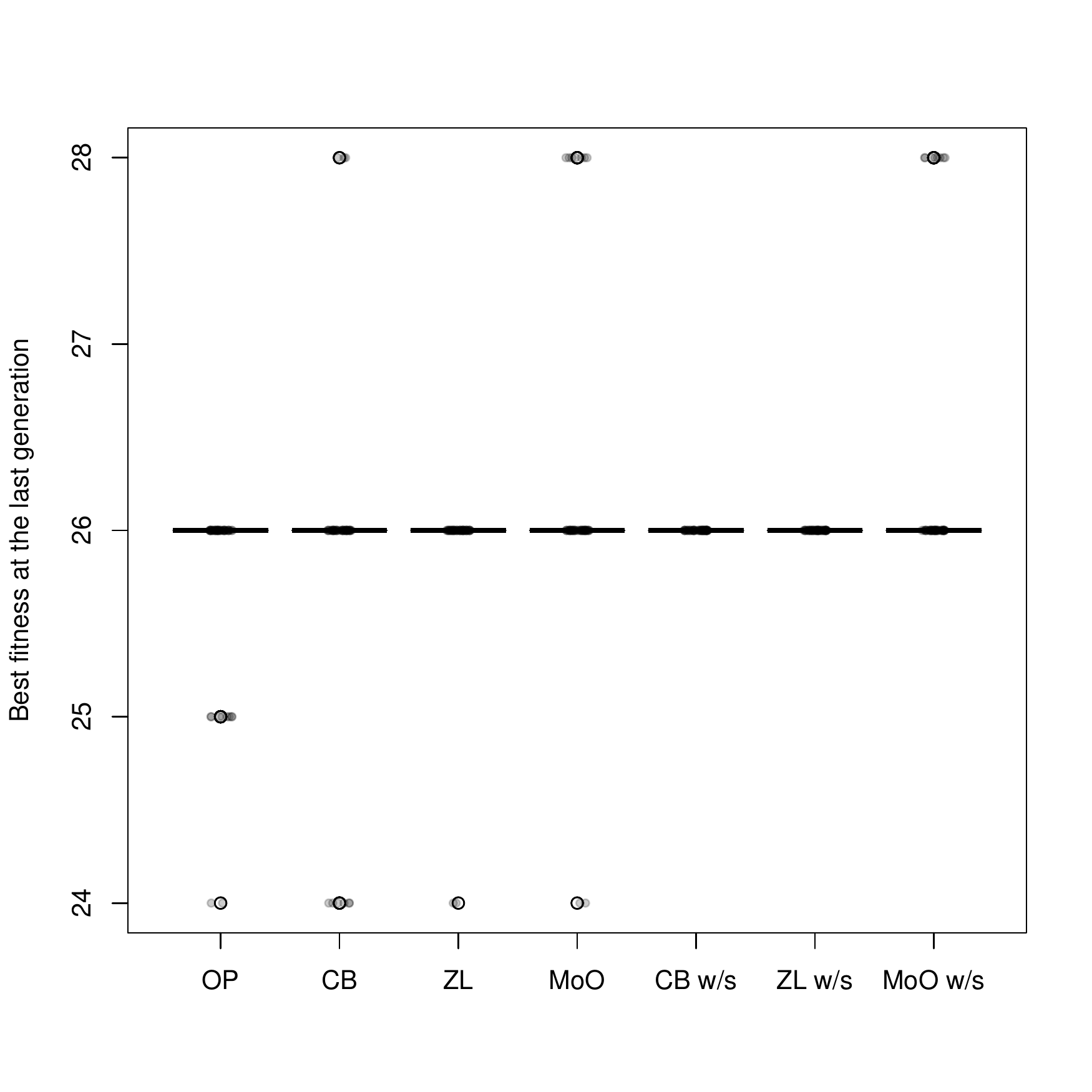}
  \caption{Results of the bent functions problem, $n=6$ instance.}
  \label{fig:bent-nvar6}
  \includegraphics[scale=0.5]{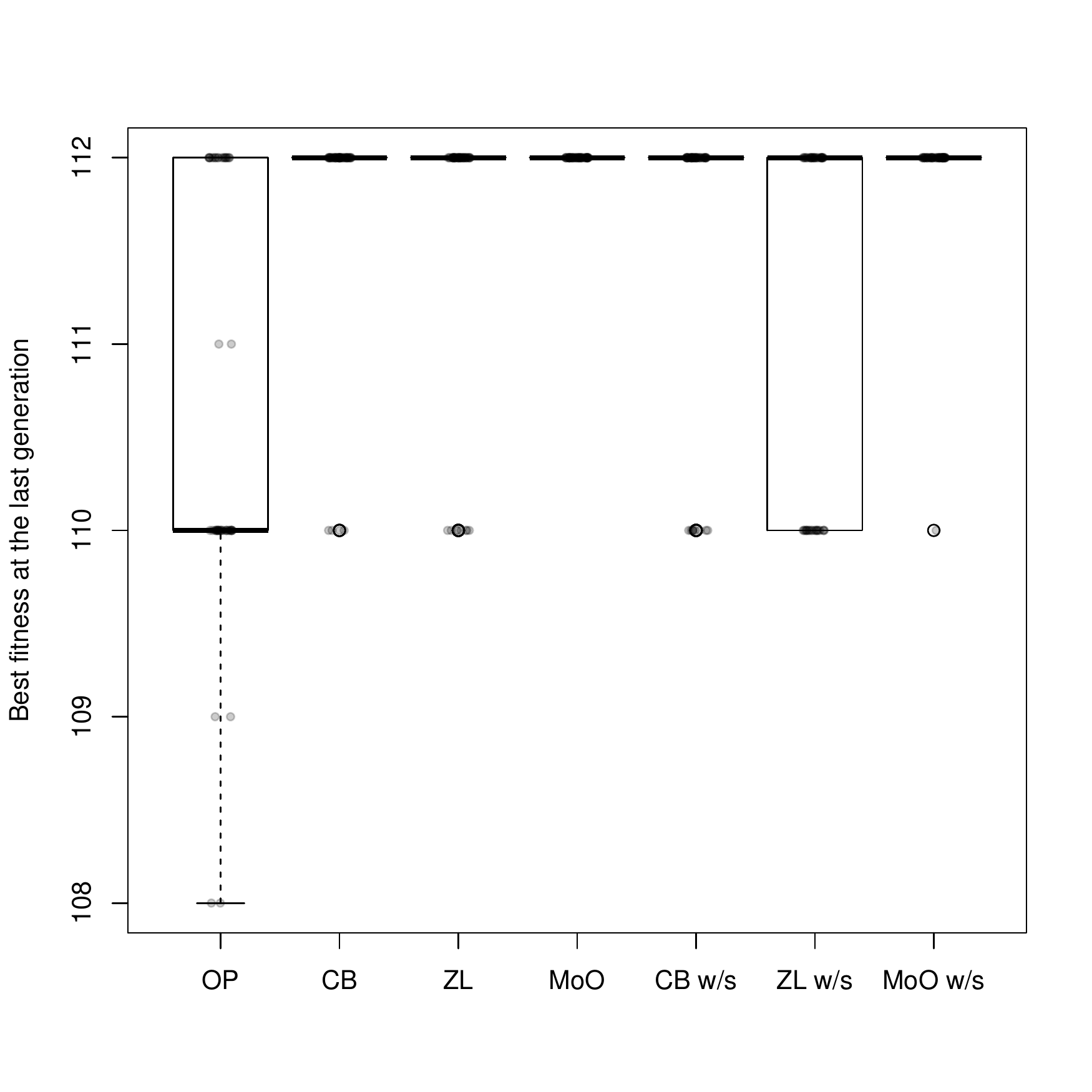}
  \caption{Results of the bent functions problem, $n=8$ instance.}
  \label{fig:bent-nvar8}
\end{figure}

\begin{figure}
  \centering
  \includegraphics[scale=0.5]{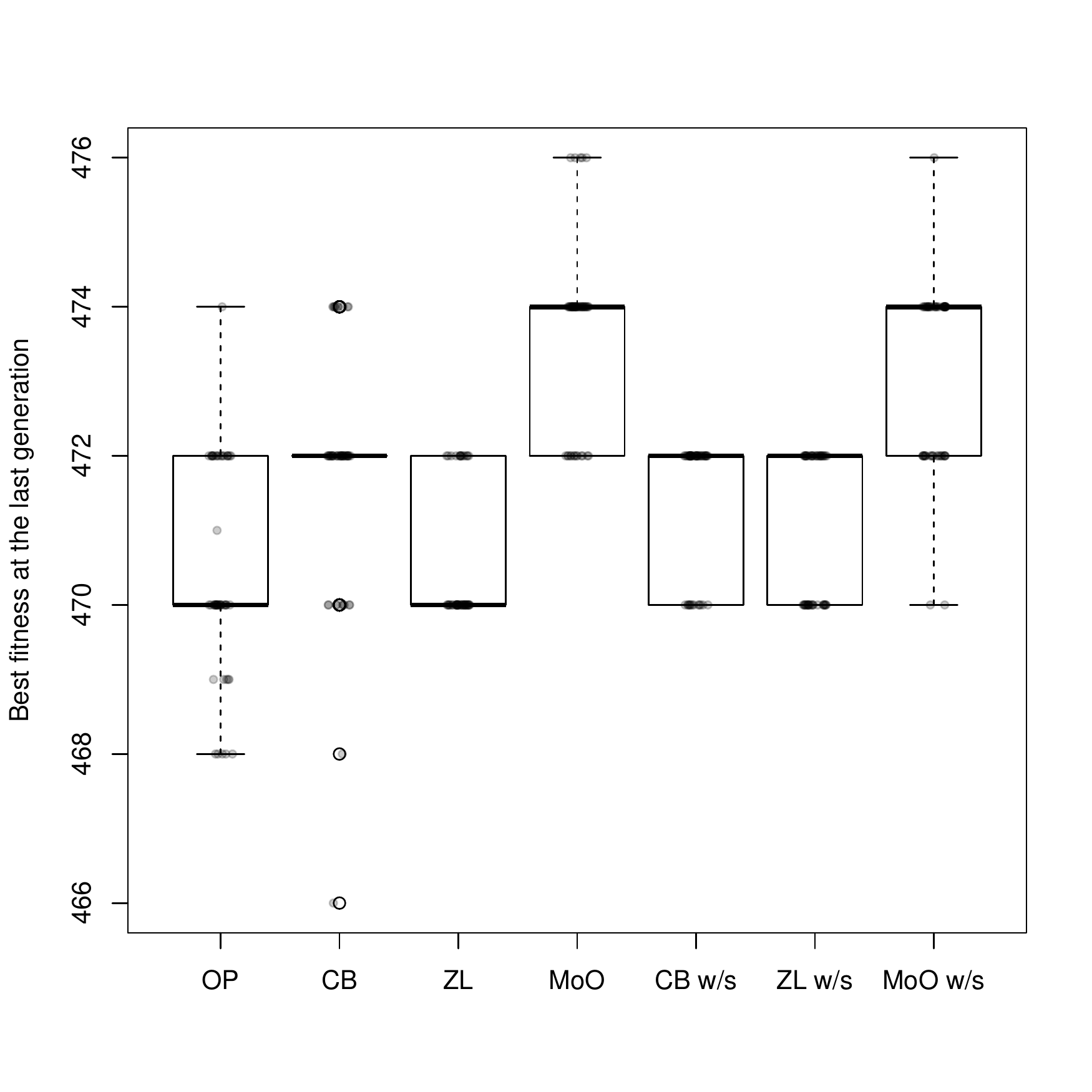}
  \caption{Results of the bent functions problem, $n=10$ instance.}
  \label{fig:bent-nvar10}
\end{figure}

Figures~\ref{fig:bent-nvar6}--~\ref{fig:bent-nvar10} report the results for the
bent functions problem.

\label{subsec:res-bent}

For the $n=6$ instance there is no clear distinction among the boxplots of the
different crossover operators, although it can be seen that balanced operators
(in particular, left-to-right counter-based and both versions of map-of-ones)
are the ones reaching the optimal nonlinearity value of $28$. We observed
moreover that one-point crossover performs worse in the statistical tests than
the left-to-right version of map-of-ones and the shuffling versions of all
balanced operators ($p$-values respectively of $0.0008$, $0.0008$, $0.0009$ and
$2.45 \cdot 10^{-5}$). Similarly to the results obtained for the balanced
Boolean functions problem, the situation changes for the $n=8$ instance. In this
case, OP performs worse than all other crossover operators, again with the
exception of the shuffling version of the zero-length crossover operator, where
no statistically significant difference arose. This observation is also
supported for $n=10$ variables, where there are no significant differences
between OP and both versions of ZL. On the other hand, OP performs worse than
both versions of CB and MoO ($p$-values respectively of $9.98\cdot 10^{-6}$,
$1.93 \cdot 10^{-15}$, $7.56 \cdot 10^{-5}$ and $1.14 \cdot 10^{-12}$).

Concerning the balanced operators, we observed no statistically significant
differences between their left-to-right and shuffling versions, except for the
zero-length crossover in the $n=8$ instance. In this case, the shuffling version
fared worse than the left-to-right counterpart ($p$-value of
$0.0044$). Analogously to the previous optimization problem, the map-of-ones
crossover resulted to be the best performer, obtaining statistically significant
differences in all comparisons with the other balanced operators over all three
problem instances, except for the $n=8$ case with the left-to-right
counter-based crossover.

\subsection{Binary Orthogonal Arrays}
\label{subsec:res-oa}
Figures~\ref{fig:bent-nvar6}--~\ref{fig:bent-nvar10} report the boxplots for the
results obtained on the orthogonal arrays problem.
Recall from Section~\ref{subsec:oa} that this is a minimization problem. Hence,
it can be clearly seen from the boxplots that one-point crossover scored the
worse performance over all three problem instances. As a matter of fact, beside
never converging to an optimal solution, all OP quantiles are substantially
higher than those featured by the distributions of the balanced operators.

Comparing the left-to-right and shuffling versions of the balanced operators, we
can observe from the boxplots that the shuffle operation actually \emph{worsens}
the performance of the counter-based and zero-length crossover over all
instances. On the other hand, the statistical tests did not detect any
significant difference between the left-to-right and shuffling versions of the
map-of-ones operator in any problem instance.

This left us with four crossover operators, namely left-to-right CB and ZL, and
both versions of MoO. Among them, however, no significant differences have been
observed through the statistical tests, in any of the three problem instances.

\section{Discussion}
\label{sec:disc}
In this section, we discuss the main findings that can be deduced from the
results reported in the previous section, especially with respect to the
research hypotheses laid out in Section~\ref{subsec:exp-par}. Additionally, we
compare our results with those obtained by two recent works in the literature,
namely~\cite{picek17} for the balanced Boolean functions and bent functions
problems and~\cite{mariot18} for the orthogonal arrays problem.

\subsection{Insights Gained from the Results}
\label{subsec:ins}
The results presented in Sections~\ref{subsec:res-bf},~\ref{subsec:res-bent}
and~\ref{subsec:res-oa} showed that one-point crossover overall scored as the
worst performer among all tested operators. The only exceptions are represented
by the $n=6$ instances of the balanced Boolean functions and bent functions
problems, where the performance of one-point was not distinguishable from that
of the counter-based and zero-length operators, and by the $n=8$ instances,
where no significant differences between one-point and zero-length crossover
could be observed. On the other hand, over the OA problem one-point crossover is
clearly worse than any balanced operator, without even the need to resort to
statistical tests for settling the question.

Hence, the main insight that we gain from these observations is that hypothesis
$H_1$, i.e.\! one-point crossover has a worse performance than any of the three
balanced crossover operators, can be considered largely confirmed by our
experiments. Using an unconstrained crossover operator such as one-point seems
to be a bad choice especially in those optimization problems where the fitness
function heavily relies on the balancedness property of the individuals, such as
in the OA case. As a matter of fact, the main term of $\fit_3$ actually measures
the balancedness of submatrices, not only of the single columns. On the other
hand, in problems where the optimization effort is not only focused on the
balancedness property, as in the balanced Boolean functions and bent functions
cases, the advantage of balanced operators over one-point is less clear on small
problem instances. This is probably due to the sufficiently limited size of the
search space for the $n=6$ instance, which allows also one-point crossover to
converge relatively easily to optimal solutions.

The indistinguishability between one-point and zero-length crossover over the
$n=8$ and $n=10$ instances for these two problems, however, seems to indicate
that the latter is not a suitable operator for optimization problems concerning
the cryptographic properties of Boolean functions. The reason could lie in the
particular representation adopted for this operator, which encodes the sequences
of adjacent zeros composing a truth table. Possibly, highly nonlinear and bent
Boolean functions could have a ``fragmented'' truth table without long runs of
zeros, hence making this encoding less efficient for these problems. As far as
we are aware there are no works in the literature that investigate the
distribution of run lengths in the truth tables of highly nonlinear and bent
functions. Hence, this could represent an interesting direction for future
research to test the ``fragmentation'' conjecture.

The second main remark that we can obtain from our experiments concerns the
ordering bias of the ``left-to-right'' versions of our crossover
operators. Surprisingly enough, we found out not only that the shuffling
operation does not give any advantage over the Boolean functions problems with
the counter-based and zero-length crossovers, but also that shuffling actually
\emph{damages} their performances over the OA problem. Therefore, we can
reasonably assert that our experiments disprove hypothesis $H_2$, i.e.\! the
shuffling versions of the counter-based and zero-length crossover have a better
performance than their ``left-to-right'' counterparts. On the other hand, since
in our results we found no statistically significant differences between the
``left-to-right'' and shuffling versions of the map-of-ones operator, we can
consider hypothesis $H_3$ confirmed.

The reason for the worse performances scored by counter-based and zero-length
crossovers with shuffling on the OA problem could reside in the particular
structure of this optimization problem. Indeed, the fitness function $\fit_3$
tries to minimize the number of repetitions of $t$-tuples going beyond the index
$\lambda$. However, crossover is applied column-wise on the Boolean matrices
represented by the parents, and when shuffling is adopted, each column of the
child is created using a different random permutation of the positions. This
could pose a problem since if two parents have a good fitness value on a
particular subset of $t$ columns, changing the order of positions when crossing
the single column could worsen the multiplicities of the $t$-tuples, thus
creating a child with worse fitness. On the other hand, the ``left-to-right''
version does not have this problem, since it adopts always the same order to
copy the positions of the columns. More specifically, we suspect that there is
nothing special about the left-to-right ordering: the important thing is that in
the OA problem the same ordering of positions should be used over all columns,
instead of different ones. It would be interesting to investigate this idea in
future research by performing further experiments with fixed positions orderings
other than the left-to-right one.

\subsection{Comparison with other Algorithms}
\label{subsec:comp}
As noted in Section~\ref{subsec:exp-par}, we selected the parameter values for
the GA such as mutation probabilities, population sizes, numbers of fitness
evaluations and runs in order to match the experimental settings of two recent
works in the literature, namely Picek et al.'s work about immunological
algorithms for evolving cryptographic Boolean functions~\cite{picek17} and
Mariot et al.'s paper about the design of binary orthogonal arrays through
evolutionary algorithms~\cite{mariot18}. For completeness, we report in this
section a comparison of our results with these two works. However, a perfect
comparison is not feasible due to some small differences concerning both the
adopted fitness functions and the structure of the optimization
algorithms. Moreover, we recall that the main focus of this paper is not about
finding a balanced crossover operator that can outperform all other
state-of-the-art algorithms, but rather analyzing whether balanced crossover
operators give an advantage over unbalanced ones.

In Table~\ref{tab:comp-1} we present the comparison of the median fitness values
for balanced and bent Boolean functions. Among our crossover operators, we
considered only the best performing one, that is, the ``left-to-right''
map-of-ones crossover.
\begin{table}[t]
\centering
\caption{Median fitness values comparison between our GA with map-of-ones
  crossover and the algorithms used in~\cite{picek17}.}
\begin{tabular}{cccccc}
\hline\noalign{\smallskip}
Problem & Instance & MoO & GA$_b$ & CLONALG$_b$ & opt-IA$_b$ \\
\noalign{\smallskip}\hline
\noalign{\smallskip}
  \multirow{3}{*}{{\sc Bal-NL}}    & $n=6$    & $26$ & $26$ & $26$ & $26$ \\
  \noalign{\smallskip}
                                   & $n=8$    & $112$ & $112$ & $112$ & $112$ \\
  \noalign{\smallskip}\hline
  \multirow{3}{*}{{\sc Bent}}      & $n=6$     & $26$ & $26$ & $28$ & $28$  \\
  \noalign{\smallskip}
                                   & $n=8$     & $112$ & $114$ & $114$ & $114$ \\
  \noalign{\smallskip}
                                   & $n=10$    & $474$ & $478$ & $478$ & $476$ \\
\noalign{\smallskip}\hline
\noalign{\smallskip}
\end{tabular}
\label{tab:comp-1}
\end{table}
From~\cite{picek17} we considered the following optimization methods: Genetic
Algorithms (GA$_b$), Clonal Selection Algorithms (CLONALG$_b$), and Optimization
Immune Algorithms (opt-IA$_b$), all using a binary encoding of the solutions. We
discarded Evolution Strategies since they resulted to be the worst performers
among all considered methods in~\cite{picek17}, and all algorithms employing the
floating point representation, since it is out of scope for the
comparison. Regarding the balanced Boolean functions problem, we observe that
all medians coincide over all problem instances and optimization methods. This
can be interpreted as the fact that the problem can be successfully approached
in different ways through optimization algorithms whose parameters are carefully
tuned or using ad-hoc variation operators. On the other hand the map-of-ones
crossover performs slightly worse than the other considered methods on the bent
functions problem, with a more pronounced difference in the median fitness for
higher number of variables.

Table~\ref{tab:comp-2} compares the success rates for the OA problem between our
map-of-ones crossover and the Genetic Algorithms with counter-based crossover
(GA-CB) and Genetic Programming (GP) proposed in~\cite{mariot18}.
\begin{table}[t]
\centering
\caption{Success rates comparison between our GA with map-of-ones crossover and
  the algorithms used in~\cite{mariot18}.}
\begin{tabular}{cccccc}
\hline\noalign{\smallskip}
Problem & Instance & MoO & GA-CB & GP \\
\noalign{\smallskip}\hline
\noalign{\smallskip}
  \multirow{3}{*}{{\sc Bin-OA}}    & $OA(16, 8, 2, 4)$  & $50$ & $27$ & $100$  \\
  \noalign{\smallskip}
                                   & $OA(16, 8, 3, 2)$  & $4$ & $3$ & $100$  \\
  \noalign{\smallskip}
                                   & $OA(16, 15, 2, 4)$ & $0$ & $0$ & $93$ \\
\noalign{\smallskip}\hline
\noalign{\smallskip}
\end{tabular}
\label{tab:comp-2}
\end{table}
One can remark that over the smallest problem instance the map-of-ones crossover
converges more frequently than the GA-CB algorithm. In the other instances,
however, both algorithms have difficulties in generating optimal solutions over
all considered runs. In particular, neither GA-CB nor the map-of-ones crossover
can match the performances of GP, which achieves a full success rates in two out
of three instances, while it converges 93\% of the time on the largest
instance. Thus, the difference in the performances between GA and GP does not
seem to be due to the particular variation operator employed, but rather to
something about the GP representation of the candidate solutions.

\section{Conclusions}
\label{sec:conc}
In this paper, we investigated the effect of three balanced crossover operators
in constraining the search space explored by a steady state GA over three
combinatorial optimization problems from the domains of cryptography and
combinatorial designs. The considered operators were the counter-based,
zero-length, and map-of-ones crossovers and all of them were studied in both in
their ``left-to-right'' and shuffled variants.

Regarding the novelty of the crossover operators, the counter-based crossover
operator is a slightly modified version of the crossover designed by Millan et
al.~\cite{millan98}, while to the best of our knowledge the zero-lengths and map
of ones crossovers are proposed for the first time in the present work.

We explored three different research questions: whether there is an advantage in
using balanced crossover operators as opposed to one-point crossover, whether
the shuffled versions of the counter-based and zero-length crossovers perform
better than their ``left-to-right'' versions, and whether the shuffled version
of the map-of-ones crossover performs as its ``left-to-right'' version.

The Mann-Whitney-Wilcoxon test was used to compare the performances of these
balanced operators with that of one-point crossover, which does not enforce any
constraint on the Hamming weight of the bitstrings. The obtained results showed
that, in general, the balanced crossovers perform better than the one-point
crossover, showing the suitability of the proposed operators for constraining
the search space.

We also noticed that using shuffling either produces no significant difference
or actually damages the performances in the case of the counter-based and
zero-length crossover, showing that the their ``left-to-right'' versions might
actually be more suitable. However, for the map-of-ones crossover the use of
shuffling neither hinder nor improve the search.

Multiple research avenues remain open for future research. The proposed balanced
crossover operators, while better than one-point crossover, does not obtain
state of the art performances in tackling combinatorial optimization problems
related to cryptography and combinatorial design theory. Further improving and
tuning our crossover operators is then an essential step in ensuring their
practical applicability in the future. Moreover, it would be interesting to
discover the effect on the shape of the fitness landscape induced by the
different balanced crossover operators proposed in this paper.

\bibliographystyle{plain}
\bibliography{bibliography}
	
\end{document}